\documentclass{article}

\usepackage[margin=1in]{geometry}

\usepackage{graphicx}
\usepackage{adjustbox}
\setkeys{Gin}{width=\linewidth,height=\textheight,keepaspectratio}

\usepackage{mathptmx} % Times New Roman 字体
\usepackage{amsmath}
\usepackage{hyperref}
\usepackage{caption}
\usepackage{float}
\usepackage{cite}
\usepackage{etoolbox}
\usepackage{booktabs}
\usepackage{longtable}
\usepackage{textcomp}
\usepackage{tabularx}
\usepackage{array}
\usepackage{listings}
\usepackage{xcolor}

\graphicspath{{figures/}}

\begin{document}

% 标题、作者、摘要
\title{ANX: Protocol-First Design for AI Agent Interaction with a Supporting 3EX Decoupled Architecture}
\author{Xu Mingze \\ Hangzhou Ziyou Data Technology Co., Ltd., Hangzhou 310000, China \\ mountcoding@gmail.com}
\date{April 4th, 2026}
\maketitle

\begin{abstract}
AI agents, autonomous digital actors, need agent-native protocols; existing methods include GUI automation and MCP-based skills, with defects of high token consumption, fragmented interaction, inadequate security, due to lacking a unified top-level framework and key components, each independent module flawed. To address these issues, we present ANX, an open, extensible, verifiable agent-native protocol and top-level framework integrating CLI, Skill, MCP, resolving pain points via protocol innovation, architectural optimization and tool supplementation. Its four core innovations: 1) Agent-native design (ANX Config, Markup, CLI) with high information density, flexibility and strong adaptability to reduce tokens and eliminate inconsistencies; 2) Human-agent interaction combining Skill’s flexibility for dual rendering as agent-executable instructions and human-readable UI; 3) MCP-supported on-demand lightweight apps without pre-registration; 4) ANX Markup-enabled machine-executable SOPs eliminating ambiguity for reliable long-horizon tasks and multi-agent collaboration. As the first in a series, we focus on ANX’s design, present its 3EX decoupled architecture with ANXHub and preliminary feasibility analysis and experimental validation. ANX ensures native security: LLM-bypassed UI-to-Core communication keeps sensitive data out of agent context; human-only confirmation prevents automated misuse. Form-filling experiments with Qwen3.5-plus/GPT-4o show ANX reduces tokens by 47.3\% (Qwen3.5-plus) and 55.6\% (GPT-4o) vs MCP-based skills, 57.1\% (Qwen3.5-plus) and 66.3\% (GPT-4o) vs GUI automation, and shortens execution time by 58.1\% and 57.7\% vs MCP-based skills.
\end{abstract}


% 正文章节
\section{Introduction}

\subsection{Background and Problem Statement}

AI agents are increasingly emerging as autonomous, goal-driven entities capable of performing complex digital tasks on behalf of humans, positioning them as the next generation of core internet users. This paradigm shift demands agent-native interaction frameworks rather than the traditional human-centric paradigms. However, existing approaches remain fragmented without a unified top-level architecture to orchestrate them into a coherent ecosystem—a critical research gap addressed by the ANX (AI Native eX) protocol: a holistic agent-native framework designed to govern, unify, and elevate GUI automation, MCP, skill systems, and CLI execution under a single systematic paradigm.

Graphical user interfaces (GUIs), human-designed by nature, incur exorbitant token costs, poor scalability, and inherent security flaws when automated for agents. MCP and skill-based CLIs offer incremental improvements but suffer from core limitations: pre-installation requirements, no use-and-go capability, separated human-agent interaction channels, overreliance on limited Markdown, and the lack of unified structured formats and sensitive data control mechanisms. Recent CLI-based explorations such as OpenCLI (Kasibatla et al., 2025) and SkillWeaver (Zheng et al., 2025) still remain browser-centric, plagued by DOM fragility, parsing barriers from page encryption, update-induced latency, and inconvenient device-bound sensitive data entry, compromising both usability and security.

Natural language ambiguity further undermines the stability of LLM-based execution in complex standard operating procedures (SOPs). Chain-of-Thought (Wei et al., 2022) fails to resolve this ambiguity, while Chain-of-Verification (Dhuliawala et al., 2024) reduces hallucinations only at the cost of additional token overhead, without eliminating reliance on unstructured natural language. Although AI agents outperform humans in routine task execution (Wang et al., 2025), their purely programmatic approaches lack quality and fail to support long-horizon, multi-step SOP scheduling and multi-agent collaboration—key capabilities for practical deployment.

Most critically, security is not treated as a first-class concern in existing paradigms. Sensitive data (e.g., passwords, financial information) must traverse the agent’s context, creating inherent privacy risks and leakage vulnerabilities, as demonstrated by Rosenberg et al. (2025) in banking agent hallucination incidents. While recent optimizations like semantic tool discovery (Mudunuri et al., 2026) and CLI-based interfaces (Gao et al., 2026) boost efficiency, they remain piecemeal solutions that cannot simultaneously address the core challenges of efficiency, security, and semantic precision for agent interaction.

ANX fills this void as a higher-order architectural paradigm that integrates, coordinates, and transcends these isolated tools. Treating GUI, MCP, CLI, and skill systems as functional execution units, ANX acts as a unifying governance framework that resolves their individual limitations, enabling systematic, secure, and efficient large-scale agent interaction.

\subsection{Contributions}

This paper presents ANX (AI Native eX) -- a co‑designed protocol and architecture for agent‑native interaction. The contributions are organized into two categories: protocol‑level innovations and architecture‑level innovations.

\textbf{Protocol Innovations (ANX Protocol)}

\begin{itemize}
\item
  \textbf{Agent-Native Design (ANX Config, ANX Markup, ANX CLI)~--} Designed first for agents: ANX provides a unified, compatible, and extensible protocol that eliminates interface description inconsistency across platforms. ANX Markup is a compact structured encoding, more concise than natural language or plain Markdown, offering higher information density, significantly reducing token consumption, and improving processing efficiency. It is easy for agents to understand and operate, with elegant, composable mechanisms, rich expressiveness, and abundant built-in utilities to support diverse complex applications.
\item
  \textbf{Human-Agent Shared Interaction}~-- The same ANX definition can be directly rendered as a usable human interface within dialog boxes (e.g., embedded UI), fundamentally extending human-agent interaction paradigms. It can also be rendered as H5 or mini\textbf{-}program, serving as a conventional application. This enables a single protocol to seamlessly support both agent-driven and human-driven interactions -- develop once, use everywhere, eliminating fragmentation and reducing development costs.
\item
  \textbf{Create-on-Demand, Use-and-Go}~-- Agents autonomously generate new ANX applications on-demand: concise, fast, dynamically parsed and hot-started. Functions load on-demand, lightweight with no download. Create, use, discard -- no pre-registration. Zhang et al. (2025) systematically compare API‑based and GUI‑based LLM agents, revealing significant differences in architecture complexity, development workflow, and user interaction models. In contrast, ANX eliminates this complexity: once an application is defined in ANX format, the 3EX architecture automatically adapts it for agent use, requiring no intricate structure design or debugging.
\item
  \textbf{Clear and Machine-Executable SOP Expression}~-- ANX's structured semantics (ANX Markup) eliminate natural language ambiguity, making standard operating procedures (SOPs) unambiguous and machine-executable. This enables reliable transmission and execution of complex workflows, long\textbf{-}horizon task scheduling.
\end{itemize}

\textbf{Architecture Innovations (3EX Decoupled Architecture)}

\begin{itemize}
\item
  \textbf{3EX (Expression-Exchange-Execution) Layered Architecture with ANXHub~}-- A general-purpose architecture designed for agents. It separates task specification (Expression), tool discovery and negotiation (Exchange), and execution (Execution) into independent, composable layers. This decoupling significantly reduces redundant steps, lowering the operational difficulty for agents -- each layer has a clear responsibility, freeing agents from handling mixed concerns. In complex scenarios, 3EX achieves higher efficiency and lower cognitive load for agents. The architecture realizes the above protocol features.~Unlike MCP's local or directory-based tool indices, ANXHub is a massive application marketplace:~it enables dynamic discovery of a vast ecosystem of ANX applications on-demand, with no pre-installation, no local binaries, and true use-and-go, eliminating virus risks at the source.
\item
  \textbf{Embedded Security via Direct UI-to-Core Communication}~-- ANX UI renders interactive forms directly in the user dialog that communicate directly with ANX Core, bypassing the LLM entirely. Sensitive data entered by the user never enters the agent's context, achieving true data isolation. Human confirmation dialogs are rendered in ANX UI format and can only be completed by human action -- the execution layer is designed such that confirmation operations cannot be delegated to the agent, eliminating the risk of automated misuse.
\item
  \textbf{Stable Long-Horizon SOP Execution and Multi‑Agent \& Human Collaboration --} Based on the 3EX architecture and ANX Markup's structured semantics, ANX enables reliable scheduling of complex workflows. Multiple agents and humans can collaborate on a single SOP -- each agent handles subtasks, while humans act as gatekeepers at critical steps (e.g., blocking approval), with consistent state managed by ANX Nodes and full security inheritance (human‑only confirmation).
\end{itemize}

Summary: By integrating these protocol and architecture innovations, ANX provides the first holistic solution that simultaneously addresses efficiency, security, stability, and multi\textbf{-}agent \& human collaboration for agent‑native interaction.

\subsection{Paper Structure}

Section 2 reviews related work; Section 3 presents ANX architecture; Section 4 describes experimental methodology; Section 5 reports results and discussion; Section 6 concludes and outlines future directions.
\section{Related Work: Four Dimensions of Agent Interaction Paradigms}

We categorize existing agent interaction approaches into four streams: GUI automation, tool‑calling protocols, emerging agent‑native protocols, and multi‑agent collaboration frameworks. To systematically assess their capabilities and limitations, we propose a four‑dimensional evaluation framework: \textbf{Protocol \& Tooling}, \textbf{Discovery \& Retrieval}, \textbf{Security}, and \textbf{Collaboration} (SOP \& Multi‑Agent). This analysis reveals critical gaps that motivate the design of ANX, which we present in Section~\ref{sec:anx-architecture}.

\subsection{GUI-Based Agent Approaches}
GUI‑based agents simulate human operations through visual recognition (e.g., screen parsing, element detection). 
\textit{Strength:} No website modification needed; highest generality. 
\textit{Limitation:} Weak spatial‑semantic alignment, high token cost, brittle to DOM changes, no security.

\subsection{Tool-Calling Protocols: MCP and Extensions}
The Model Context Protocol (MCP) (Ding et al., 2025) standardizes tool invocation via JSON schemas. 
\textit{Strength:} Declarative, composable interface. 
\textit{Limitation:} Pre‑installation required, token overhead for dynamic options, no data isolation. Extensions like semantic tool discovery (Mudunuri et al., 2026) improve retrieval but remain tool‑centric.

\subsection{Emerging Agent-Native Protocols}
\begin{itemize}
\item \textbf{CHEQ} (Rosenberg et al., 2025): human‑in‑the‑loop confirmation. \textit{Strength:} standardised confirmation. \textit{Limitation:} no data isolation, agent sees sensitive data before confirmation.
\item \textbf{VACP} (Stähle et al., 2026): visual analytics context compression. \textit{Strength:} reduces token consumption. \textit{Limitation:} domain‑specific.
\item \textbf{FormFactory} (Li et al., 2025): benchmark for form filling. \textit{Limitation:} no protocol or architecture.
\end{itemize}
None offer a holistic solution.

\subsection{Agent Architecture and Governance}
Existing work (Ben Hassouna et al., 2026; Chen et al., 2025; Dochkina, 2026) focuses on agent internals (reasoning, memory, self‑organization) rather than interaction paradigms, leaving protocol‑level inefficiencies and security gaps unaddressed.

\subsection{Four-Dimensional Evaluation Framework}

To enable a fair, multidimensional comparison, we define four orthogonal dimensions:

\begin{enumerate}
\item \textbf{Protocol \& Tooling}: task representation, tool discovery, runtime environment, robustness, token efficiency, use‑and‑go, shared specification, architectural decoupling.
\item \textbf{Discovery \& Retrieval}: discovery mechanism, scope, pre‑installation, latency, dynamic updates, composability.
\item \textbf{Security}: data isolation (LLM visibility), human confirmation, authentication, threat model, workflow integration.
\item \textbf{Collaboration (SOP \& Multi‑Agent)}: SOP representation, determinism, dependencies/branching, multi‑agent coordination, human‑in‑the‑loop, scale, auditability, tool integration.
\end{enumerate}

\subsubsection{Protocol \& Tooling Dimension}

\begin{table}[H]
  \centering
  \caption{Protocol \& Tooling Dimension: Comparison of Representative Approaches}
  \label{tab:tool-compare}
  \small
  \begin{tabularx}{\linewidth}{@{}lXXXXX@{}}
  \toprule
  Dimension & GUI Automation & MCP & OpenCLI & SkillWeaver & ANX \\
  \midrule
  Task Representation & NL+vision & JSON schema & CLI commands & Reusable APIs & Structured ANX Markup \\
  Tool Discovery & None (DOM) & Pre‑registered & Pre‑converted catalog & Autonomous exploration & Dynamic semantic (ANXHub) \\
  Runtime Env. & Browser+DOM & Tool server & Browser+CLI & Browser+DOM & ANX Core+CLI (no browser) \\
  Robustness & Low (brittle) & N/A & Low (DOM‑dep) & Low (brittle UI) & High (no DOM) \\
  Use‑and‑Go & Yes & No & Yes (catalog) & No & Yes (dynamic) \\
  Token Efficiency & Low & Medium & High & Medium & High\\
  Shared Spec & For Human & No & For Agent & For Agent & Yes (native) \\
  Decoupling & Monolithic & Tool‑centric & Browser‑centric & Environment‑bound & 3EX layered \\
  \bottomrule
  \end{tabularx}
\end{table}

\textbf{Discussion and Summary:} The protocol \& tooling dimension reveals a clear trade‑off. GUI automation offers maximum generality but suffers from high token consumption, DOM brittleness, and no built‑in security. MCP provides a declarative tool interface but requires pre‑installation and lacks dynamic discovery. OpenCLI enables zero‑install CLI usage but remains browser‑dependent. SkillWeaver allows autonomous skill discovery but relies on fragile UI sequences. ANX uniquely combines a compact structured task representation (ANX Markup), dynamic semantic discovery (ANXHub), a browser‑free lightweight runtime (ANX Core+CLI), high efficiency through progressive disclosure, high‑density protocol, and layered architecture, native shared human‑agent specification, and full architectural decoupling (3EX). These features together address the fragmentation and inefficiency that plague existing approaches.

\subsubsection{Discovery \& Retrieval Dimension}

\begin{table}[H]
  \centering
  \caption{Discovery \& Retrieval Dimension: Comparison of Representative Approaches}
  \label{tab:discovery-compare}
  \small
  \begin{tabularx}{\linewidth}{@{}lXXXX@{}}
  \toprule
  Dimension & Semantic Tool Discovery & Skill Discovery & MCP (Basic) & ANX \\
  \midrule
  Discovery Mechanism & Vector retrieval & Autonomous exploration & Pre‑registered list & Semantic vector (ANXHub) \\
  Discovery Scope & Local tools & Current environment & Local tools & Global marketplace \\
  Pre‑installation Required & Yes & No (env‑dep) & Yes & No (zero‑install) \\
  Discovery Latency & Low & High & Very low & Medium (optimisable) \\
  Dynamic Updates & Re‑indexing & Re‑exploration & Re‑registration & Real‑time publish \\
  Applicable Scenarios & Fixed toolset & New environment & Static toolset & Open, dynamic \\
  Composability & Low & Medium & Low & High (SOP composition) \\
  \bottomrule
  \end{tabularx}
\end{table}

\textbf{Discussion and Summary:} Existing discovery mechanisms are either static (MCP's pre‑registered list), limited to local tools (semantic vector retrieval), or environment‑specific and costly (skill discovery). None offer a global, real‑time, zero‑install marketplace. ANX introduces a semantic vector retrieval layer (ANXHub) that provides on‑demand discovery from a massive application marketplace, with zero pre‑installation, real‑time updates, and high composability through SOP composition. While discovery latency is currently medium, it is optimisable; the key advantage is the ability to handle open, dynamic tasks across domains without prior registration.

\subsubsection{Security Dimension}

\begin{table}[H]
  \centering
  \caption{Security Dimension: Comparison of Representative Approaches}
  \label{tab:security-compare}
  \small
  \begin{tabularx}{\linewidth}{@{}lXXXX@{}}
  \toprule
  Dimension & CHEQ & AIP & AgentCrypt & ANX \\
  \midrule
  Data Isolation (LLM sees?) & Yes (data already seen) & N/A (network) & No (encrypted, but LLM sees) & No (UI‑to‑Core, LLM never sees) \\
  Human Confirmation & Dialog (bypassable?) & No & No & Human‑only, no exit \\
  Authentication & None & DID‑based & End‑to‑end encryption & User Token + trusted Core/Hub \\
  Workflow Integration & Standalone & Standalone & Standalone & Native integration \\
  \bottomrule
  \end{tabularx}
\end{table}

\textbf{Discussion and Summary:} Security is largely overlooked in existing agent protocols. CHEQ provides a confirmation dialog but does not prevent the agent from seeing sensitive data beforehand. AIP and AgentCrypt focus on network‑layer or transport security, leaving the LLM context vulnerable. ANX introduces a fundamentally different approach: UI‑to‑Core direct communication via the chat dialog, ensuring that the LLM never sees raw sensitive data. Human confirmation is unbypassable (no programmatic exit), authentication is token‑based with a trusted Core/Hub. Moreover, security is natively integrated into the workflow rather than being a standalone layer. This makes ANX the first protocol to provide application‑level data isolation and hardened human‑in‑the‑loop guarantees.

\subsubsection{Collaboration Dimension (SOP \& Multi‑Agent)}

\begin{table}[H]
  \centering
  \caption{Collaboration Dimension: SOP Scheduling \& Multi‑Agent Coordination}
  \label{tab:collaboration-compare}
  \small
  \begin{tabularx}{\linewidth}{@{}lXXXX@{}}
  \toprule
  Dimension & COLLAB‑LLM & ANP & A2A & ANX \\
  \midrule
  SOP Representation & Role‑based task graph & Agent Description + semantic & Agent Card + workflows & SOP ANX Config \\
  Determinism & Medium (dynamic) & Medium (negotiation) & Medium (task‑based) & High (state machine) \\
  Dependencies \& Branching & Implicit (communication) & Peer discovery + negotiation & Task delegation & Explicit sources/targets, conditional \\
  Multi‑Agent Coordination & Yes ($\leq$8 agents) & Yes (P2P) & Yes (enterprise) & Yes (scalable, ANXHub sync) \\
  Human‑in‑the‑Loop & No & No & No & Yes (blocking, native) \\
  Workflow Scale & Small & Arbitrary & Medium‑large & Arbitrary + long‑horizon \\
  Auditability & Limited & Limited (logs) & Task‑level & Built‑in (ANX Nodes) \\
  Tool Integration & Separate & Separate & Separate & Seamless (3EX) \\
  \bottomrule
  \end{tabularx}
\end{table}

\textbf{Discussion and Summary:} Multi‑agent collaboration protocols (COLLAB‑LLM, ANP, A2A) provide valuable coordination mechanisms but lack several critical features: deterministic SOP execution, native human‑in‑the‑loop, integrated security, and seamless tool integration. ANX addresses these through its SOP framework built on ANX Config with explicit \texttt{sources}/\texttt{targets} dependencies, conditional routing, and a state machine that ensures high determinism. Multi‑agent coordination is handled by ANXHub (task decomposition and state synchronization) and scales beyond fixed limits. Human‑in‑the‑loop is supported natively via blocking, unbypassable confirmation dialogs. Auditability is built‑in through ANX Nodes, and tool integration is seamless thanks to the 3EX architecture. ANX thus provides a complete, production‑ready foundation for long‑horizon SOP execution and multi‑agent collaboration.

\subsection{Research Gaps and Open Challenges}

The comparative analysis reveals that no existing approach simultaneously addresses all four dimensions effectively:

\begin{enumerate}
\item \textbf{Protocol \& Tooling Gap}: Browser‑centric (fragile, high token) or tool‑centric (pre‑installation). No compact, structured task representation with dynamic zero‑install discovery.
\item \textbf{Discovery Gap}: Existing discovery limited to local tools (MCP, semantic) or costly environment exploration (skill discovery). No global, real‑time, zero‑install marketplace.
\item \textbf{Security Gap}: All surveyed methods either expose sensitive data to LLM context or lack application‑level, unbypassable human confirmation. Network security does not protect against LLM leakage.
\item \textbf{Collaboration Gap}: Multi‑agent protocols (A2A, ANP, COLLAB‑LLM) provide coordination but lack deterministic SOP execution, native human‑in‑the‑loop, and integrated security.
\end{enumerate}

\subsection{Summary of Academic Contributions and Takeaway for Future Comparisons}

This chapter makes the following contributions that can be directly cited by other researchers:

\begin{enumerate}
\item \textbf{A four‑dimensional evaluation framework} (Protocol \& Tooling, Discovery, Security, Collaboration) with explicit sub‑dimensions. Researchers can use these dimensions to systematically compare any new agent interaction system.
\item \textbf{A comparative taxonomy} categorising existing approaches into four families: browser‑centric (GUI, OpenCLI, SkillWeaver), tool‑centric (MCP, semantic discovery), specialised security (CHEQ, AIP, AgentCrypt), and multi‑agent collaboration (COLLAB‑LLM, ANP, A2A). Each family is characterised by its strengths and inherent bottlenecks.
\item \textbf{Identification of four persistent open challenges}: zero‑friction skill adoption, real‑time global discovery, privacy‑preserving execution (LLM never sees sensitive data), and deterministic long‑horizon SOP execution with integrated multi‑agent and human‑in‑the‑loop support.
\item \textbf{A clear, citation‑ready conclusion}: “To the best of our knowledge, no prior approach combines structured semantic task representation, dynamic global discovery, built‑in data isolation (LLM never sees sensitive data), unbypassable human confirmation, and native deterministic SOP orchestration under a single protocol. This study presents the ANX protocol and its 3EX decoupled architecture, designed to achieve full coverage of capabilities across all four dimensions and address the aforementioned research gaps.”
\end{enumerate}

\emph{For researchers and practitioners who wish to position their work against the state of the art, we recommend using the four‑dimensional framework and the taxonomy presented here as a reference. The tables and analysis above can be directly cited to support claims about the limitations of existing methods and the novelty of new contributions.}

\begin{table}[H]
  \centering
  \caption{Summary of All Approaches across Four Dimensions (Simplified)}
  \label{tab:all-tools-summary}
  \small
  \begin{tabularx}{\linewidth}{@{}lXXXX@{}}
  \toprule
  Method & Protocol \& Tooling & Discovery & Security & Collaboration \\
  \midrule
  GUI Automation & NL+vision; brittle; high token & None & No isolation; LLM sees & None \\
  MCP & JSON schema; pre‑install; medium token & Pre‑registered list & No isolation; OAuth & Sequential calls \\
  OpenCLI & CLI; DOM‑dep; zero‑install & Pre‑converted catalog & Device entry; no isolation & None \\
  SkillWeaver & Reusable APIs; brittle UI & Autonomous exploration; env‑dep & No isolation & None \\
  Semantic Discovery & Vector retrieval; local & Local tools; vector search & None (retrieval) & None \\
  Skill Discovery (gen.) & Env‑dep; no fixed protocol & Trial‑error; high latency & None & None \\
  CHEQ & Confirmation only & None & Dialog; agent sees data & None \\
  AIP & Network layer; no tasks & DID‑based discovery & Comm security; no LLM iso & None \\
  AgentCrypt & Encryption only & None & E2E encrypt; LLM sees plain & None \\
  COLLAB-LLM & Role messaging; no discovery & Pre‑defined roles & None & Dynamic graph; $\leq$8 agents; 89\% \\
  ANP & P2P; DID‑based & Semantic; capability neg. & DID auth; comm security & Multi‑agent discovery; no workflow \\
  A2A & Agent Card; enterprise & Agent Card metadata & External OAuth; no isolation & Task delegation; no SOP \\
  \midrule
  ANX & Structured markup; ANXHub; no browser; high efficiency; shared spec; 3EX & Global marketplace; zero‑install; real‑time; composable & UI‑to‑Core; LLM never sees; human‑only; user token; explicit threat & SOP (sources/targets); deterministic; scalable multi‑agent; human‑in‑loop; auditable \\
  \bottomrule
  \end{tabularx}
\end{table}
\section{ANX Protocol and Architecture}
\label{sec:anx-architecture}

ANX follows six core principles: decoupling, progressive disclosure, security by default, semantic precision, marketplace‑driven extensibility, and verifiability. The following sections detail the protocol and architecture that realize these principles.

\subsection{ANX Protocol: Structured Semantics}

The ANX protocol defines a family of structured, machine‑readable formats that enable precise agent‑system interaction.

\subsubsection{ANX Markup and ANX Config}

ANX Markup~is a hybrid structured encoding combining XML‑like forms. It clearly defines task metadata, field attributes, validation rules, interaction elements, and security annotations.~ANX Config~provides a unified configuration format that simplifies setup and is equally accessible to humans and agents, enabling seamless co‑use.

\textbf{ANX Config Format Example:}

\begin{lstlisting}
{
  "protocol": "ANX", "version": "1.0.0",
  "kind": "form",
  "title": "Create Job Seeker Account",
  "items": [
    {"nick": "lastName", "kind": "input"},
    {"nick": "industry", "kind": "options", "optionsSet": {
      "dataset": {"url_dataset": "http://localhost:7887/dataset/industries"},
      "valueNick": "id",
      "titleNick": "name"
    }}
  ]
}
\end{lstlisting}

Critically, when ANX Markup contains fields marked as~"type": "sensitive", the rendering engine of the user interface (ANX UI) communicates directly with ANX Core, completely bypassing the LLM. This ensures that the agent never receives raw sensitive data; only a reference token is passed to the execution layer, achieving robust isolation.

\textbf{ANX Markup Format Example:}

\begin{lstlisting}
<x form c_8193>
## Create Job Seeker Account
Join us to discover more career opportunities
<x input c_2354>**lastName:** Mingze</x>
<x options card_1675>
**industry:**
<x 0> Please select industry</x>
<x 1 it> Information Technology</x>
<x 2 finance> Finance</x>
</x>
<x button c_2326 tap="submit">
[Create Account](/submit)
</x>
</x>
\end{lstlisting}

\subsubsection{ANX CLI}

ANX CLI defines a lightweight, cross‑platform command‑line carrier that is independent of any specific user interface. The command format is:

\begin{lstlisting}
anx <cardKey> <action> params
\end{lstlisting}

For example, executing a job application form filling:

\begin{lstlisting}
anx c_8193 set_form '{"lastName":"Mingze","industry":"it"}'
\end{lstlisting}

ANX CLI eliminates the inconsistency of interface descriptions across different platforms, where each dialog system previously developed its own ad‑hoc format. It provides a universal, compatible, and extensible protocol.

\subsubsection{Dynamic Discovery Protocol}

The dynamic discovery protocol allows skills to be indexed rather than pre‑installed into the agent’s local environment. Applications can be discovered and used on‑demand without installation, providing true "use‑and‑go" capability. Progressive disclosure of task‑relevant information fundamentally avoids token explosion as the skill library scales. The protocol defines query semantics over a semantic index (realised by ANXHub in the architecture), returning only the top‑k relevant tools for the current task.

\subsubsection{ANX SOP Specification}

ANX SOP (Standard Operating Procedure) is a core application of ANX Markup for task scheduling. It adopts schema‑based ANX Markup to explicitly define SOP steps, conditional logic, operation standards, and field constraints. This resolves the natural language ambiguity that plagues traditional SOP descriptions.

\textbf{Code 3:} provides a concrete SOP specification example for resume screening

\begin{lstlisting}
{
  "protocol": "ANX", "version": "1.0.0",
  "kind": "sop", "title": "Resume Screening",
  "steps": [
    {"uuid": "s1", "start": true, "nick": "CollectAndParse", "kind": "form",
      "items": [
        {"name": "resumeSources", "kind": "input"},
        {"name": "parsedInfo", "kind": "textarea"}
      ]},
    {"uuid": "s2", "nick": "AIDecision", "kind": "condition", "sources": ["s1"],
      "case": [
        {"when": "matchingScore >=80, no risks/flaws", "targets": ["s3"]},
        {"when": "matchingScore <80, or risks/flaws exist", "targets": ["s4"]}
      ]},
    {"uuid": "s3", "nick": "HighMatchProcess"},
    {"uuid": "s4", "nick": "LowMatchProcess"}
  ]
}
\end{lstlisting}

\textbf{Control Flow Semantics.} ANX SOP uses two complementary mechanisms to define control flow: sources and targets. The sources field lists predecessor steps that must be completed before the current step can be scheduled, establishing static structural dependencies to ensure determinism, verifiability, and support for synchronisation and parallel patterns. The targets field provides a constrained set of candidate steps for dynamic routing, allowing the LLM to select appropriate branches based on runtime conditions. Together, they balance structural stability and adaptive decision‑making.

For step synchronisation with multiple sources, ANX SOP supports two joining strategies controlled by the sources\_join field:

\begin{itemize}
\item
  all (default): The step executes only when all executed sources have completed. Unexecuted sources due to conditional routing are marked as skipped and do not block the workflow.
\item
  any: The step executes as soon as any one of its sources has completed.
\end{itemize}

In case of overlapping declarations -- when a step appears in the predecessor's targets list and also lists the predecessor in its own sources -- dynamic routing via targets takes precedence. The step is executed only if selected by the LLM, rather than being automatically triggered by the source dependency.

\subsection{3EX Architecture}

The 3EX (Expression‑Exchange‑Execution) architecture is the runtime realisation of the ANX protocol. It separates task specification, tool discovery, and execution into three independent, composable layers, significantly reducing redundant steps and lowering the operational difficulty for agents.

\begin{figure}[htbp]
  \centering
  \includegraphics{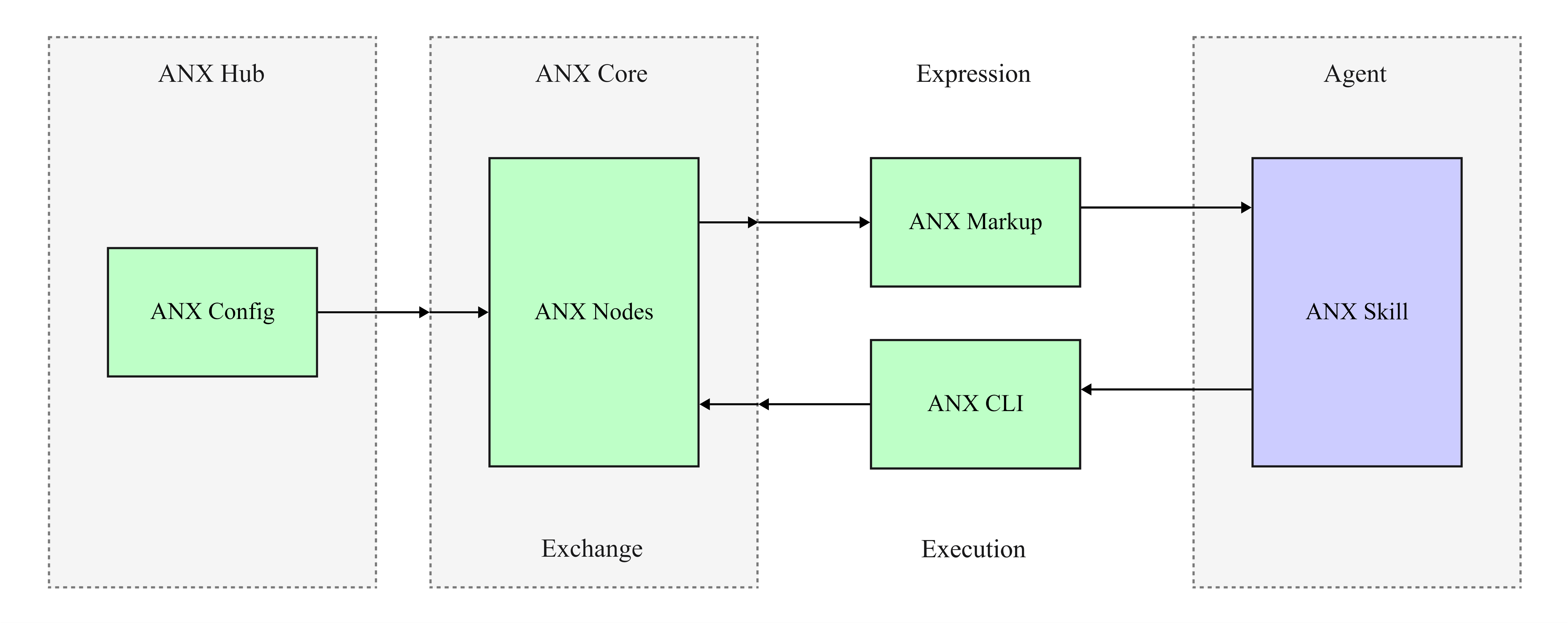}
  \caption{ANX 3EX (Expression-Exchange-Execution) Runtime Mechanism}
  \label{fig:3ex-architecture}
\end{figure}

\subsubsection{Expression Layer}

The Expression layer uses ANX Markup to represent tasks in a structured way. It holds the task metadata, field definitions, validation rules, and security annotations. This layer is responsible for producing a canonical, unambiguous description of what needs to be done.

\subsubsection{Exchange Layer (ANXHub)}

The Exchange layer is implemented by ANXHub. Unlike MCP's local or directory‑based tool indices, ANXHub is a massive application marketplace that enables dynamic discovery of a vast ecosystem of ANX applications on‑demand, with no pre‑installation, no local binaries, and true use‑and‑go. ANXHub relies on semantic vector search technology to retrieve relevant skills from the skill index. This design avoids token explosion and ensures that baseline token consumption remains fixed regardless of the size of the skill library. In addition, ANXHub is responsible for task routing and cross‑agent synchronization, laying the foundation for multi‑agent collaborative work.

\subsubsection{Execution Layer (ANX Core, ANX CLI, ANX Node)}

The Execution layer consists of:

\begin{itemize}
\item
  ANX Core -- a parser and execution engine that converts structured task specifications and retrieved tool information into platform‑agnostic ANX CLI commands.
\item
  ANX CLI -- the lightweight command carrier.
\item
  ANX Node -- an execution container that runs the commands and streams back results and state feedback.
\end{itemize}

Progressive disclosure is adopted in command generation: only the execution commands required for the current task step are generated, avoiding unnecessary token consumption.

\subsubsection{Protocol State Machine}

The ANX protocol state machine defines the complete lifecycle of an interaction. It highlights two security‑critical states:

\begin{itemize}
\item
  WAITING\_UI -- Sensitive data fields trigger this state; the UI component communicates directly with ANX Core, bypassing the LLM. The agent never sees the raw input.
\item
  CONFIRMING -- Critical actions require explicit human confirmation. This state has no programmatic exit -- completion requires human interaction, enforcing the security property that sensitive operations cannot be automated.
\end{itemize}

\begin{figure}[htbp]
  \centering
  \includegraphics{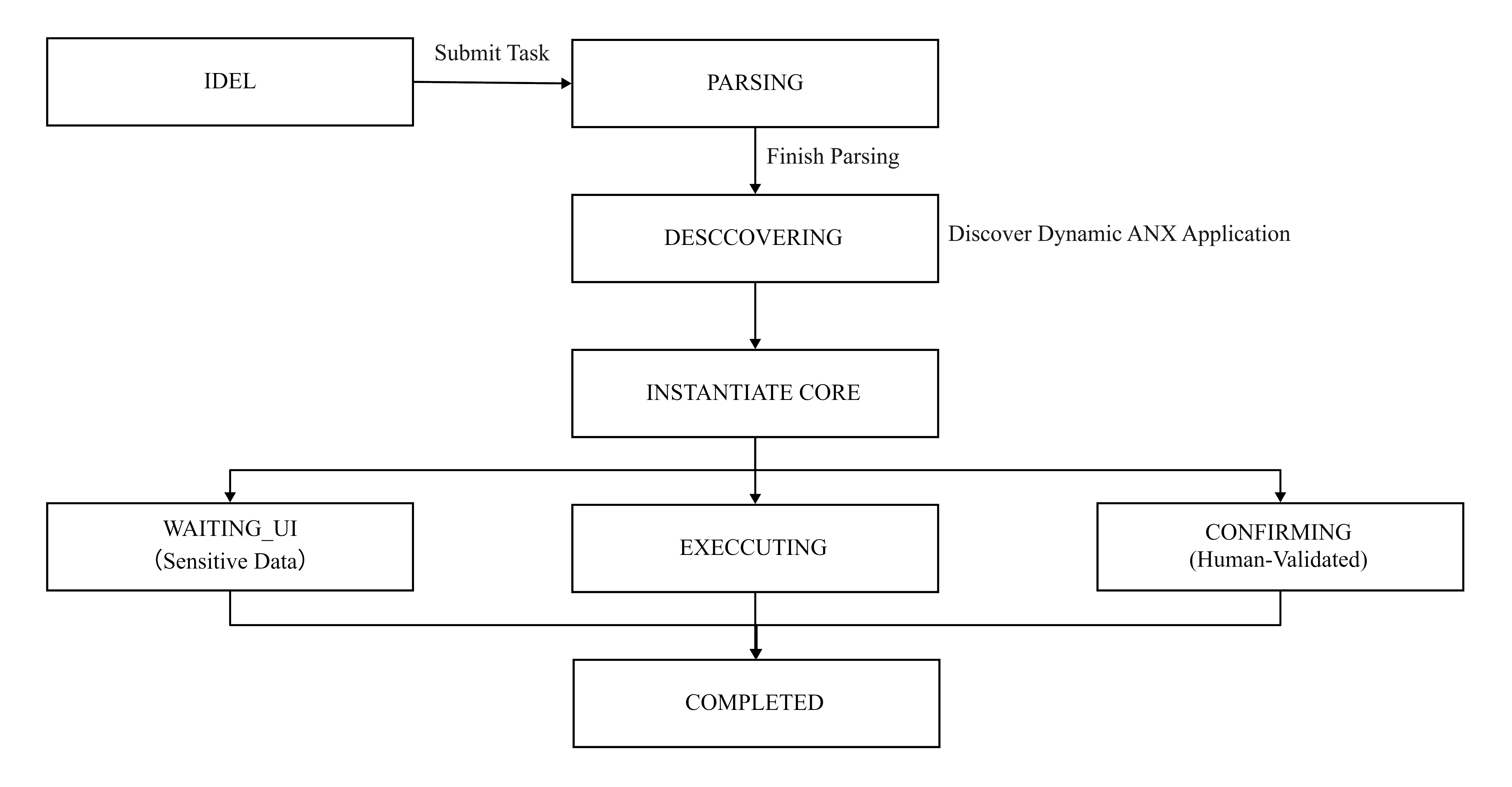}
  \caption{ANX Protocol State Machine}
  \label{fig:state-machine}
\end{figure}

\subsubsection{Key Components and Interaction Flow}

\begin{table}[htbp]
  \centering
  \caption{Main components of ANX}
  \label{tab:components}
  \fontsize{9}{11}\selectfont
  \begin{tabularx}{\linewidth}{@{}p{0.13\linewidth}p{0.80\linewidth}@{}}
  \toprule
  Component & Function \\
  \midrule
  ANXHub & Semantic tool discovery, task routing, cross-agent sync \\
  ANX Markup & Structured task representation; triggers direct UI-to-Core communication for sensitive fields \\
  ANX Core & Parser, execution engine, CLI generation \\
  ANX CLI & Lightweight command-line carrier \\
  ANX Node & Execution node with state feedback \\
  ANX Config & Global configuration for application or multi-agent \& human deployments \\
  ANX UI & Rendering engine that displays forms and confirmation dialogs, communicating directly with ANX Core \\
  \bottomrule
  \end{tabularx}
\end{table}

\textbf{Interaction Flow}

\begin{itemize}
\item
  Task Expression -- Agent or human constructs ANX Markup.
\item
  Tool Discovery -- ANXHub performs semantic search over the marketplace and returns top‑k tools.
\item
  Command Generation -- ANX Core generates ANX CLI commands.
\item
  Secure Data Handling -- Sensitive fields render as embedded forms via ANX UI; the UI communicates directly with ANX Core, bypassing the LLM. The agent never sees raw input.
\item
  Human Confirmation -- Critical actions trigger ANX UI confirmation dialogs that can only be completed by human interaction; no programmatic completion interface exists.
\item
  Execution -- ANX CLI commands execute on ANX Nodes; results stream back.
\item
  Feedback -- Agent receives structured results.
\end{itemize}

\subsection{Security Mechanisms}

ANX ensures secure agent interaction through two core, non‑overlapping mechanisms embedded in its 3EX architecture and state machine: sensitive data isolation and human validation of critical operations. Future extensions will support low‑sensitivity automation via encrypted Core storage and CLI requests, while high‑sensitivity data and critical confirmations remain human‑exclusive.

\subsubsection{UI‑to‑Core Sensitive Data Isolation}

ANX achieves strict sensitive data isolation by architectural design: when ANX Markup marks fields as sensitive, ANX UI generates dedicated forms that transmit sensitive information to ANX Core via encrypted communication, effectively avoiding the risk of intermediate data being hijacked by the LLM. Additionally, when the LLM requests to view ANX Markup from ANX Core, the Core automatically shields all sensitive information in the markup, ensuring the LLM only accesses non‑sensitive content. The agent receives only a non‑sensitive reference token, with encrypted transmission further enhancing data confidentiality.

\begin{figure}[htbp]
  \centering
  \includegraphics{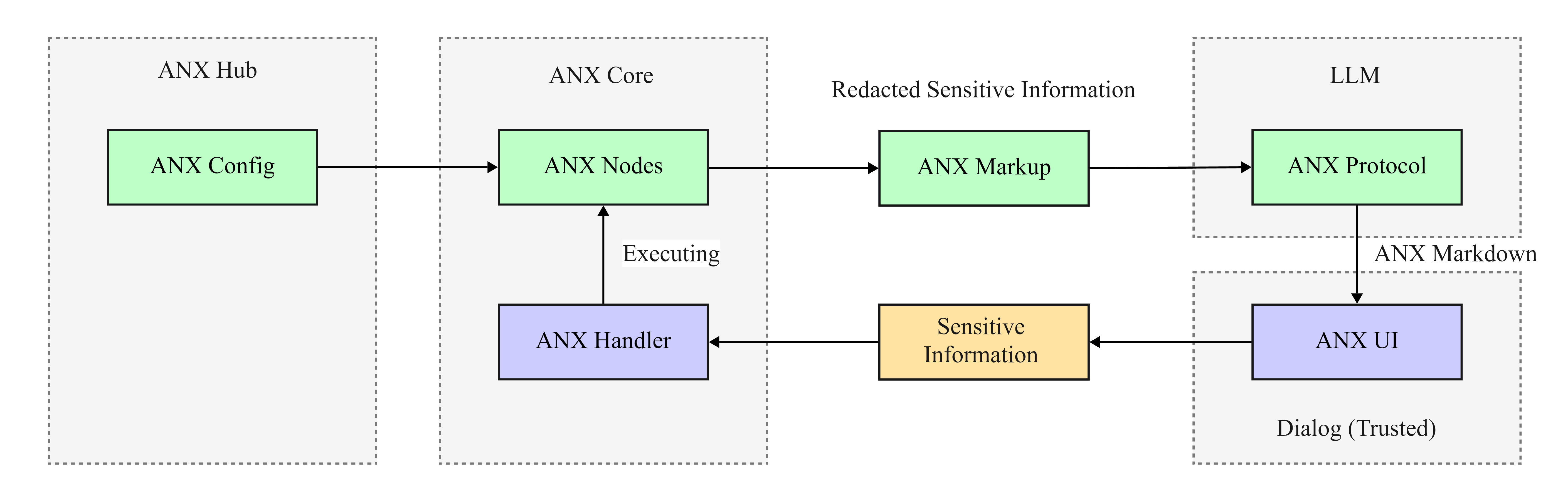}
  \caption{ANX Sensitive Information Handling Mechanism}
  \label{fig:sensitive-data}
\end{figure}

\subsubsection{Human‑Only Confirmation}

Building on the sensitive data isolation mechanism, ANX further ensures the security of sensitive operations by supporting the designation of human‑exclusive completion. To prevent the LLM from simulating human operations and unauthorised execution, ANX constructs a trusted dialog box that connects directly to ANXHub to obtain a User Token -- this User Token is exclusively managed by ANX UI and remains completely invisible to the LLM. When a Human‑Validated Request is initiated for sensitive operations, it must carry the valid User Token; only after ANX Handler verifies the legitimacy of the User Token will the sensitive operation be executed. This mechanism is integrated into the protocol's CONFIRMING state (with no programmatic exit), ensuring no agent‑accessible bypass exists and users retain absolute control over high‑risk actions.

\begin{figure}[htbp]
  \centering
  \includegraphics{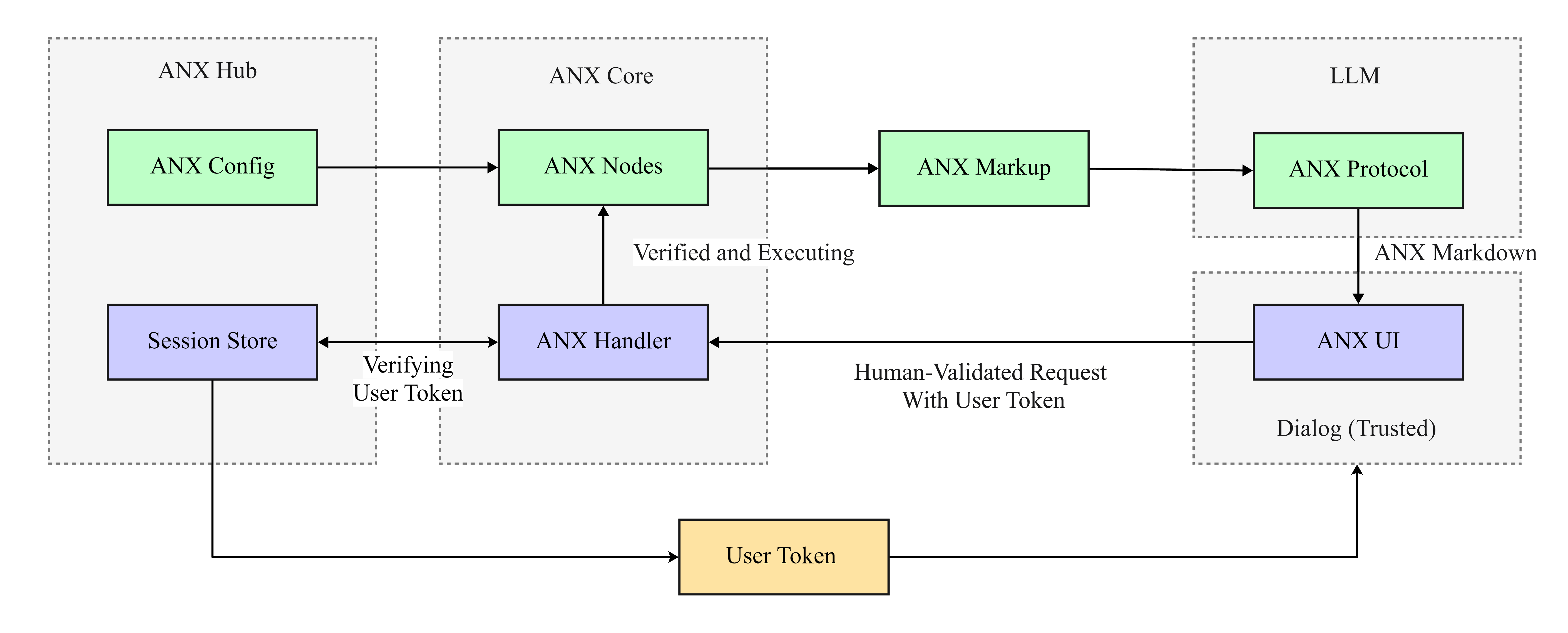}
  \caption{ANX Human-Validated Handling Mechanism}
  \label{fig:human-validation}
\end{figure}

\subsubsection{Threat Model}

ANX's built‑in security mechanisms (direct UI‑to‑Core communication, human‑only confirmation) mitigate core privacy and automation risks, but its defence boundaries must be clarified. ANX does not defend against the following scenarios:

\begin{itemize}
\item
  \textbf{Unvalidated dialog credibility} -- If users fail to distinguish legitimate ANX UI dialogs from malicious imitations, sensitive operations may be misauthorised.
\item
  \textbf{Untrusted ANX Core} -- Using a tampered or untrusted ANX Core compromises UI‑Core communication, leading to sensitive data leakage or unauthorised execution.
\item
  \textbf{Non‑official ANX Hub} -- Connecting to an untrusted ANX Hub risks malicious tool injection, task misrouting, or semantic information leakage.
\item
  \textbf{LLM‑induced social engineering} -- ANX cannot prevent LLMs from ignoring UI prompts and inducing users to manually send sensitive data directly to the LLM.
\end{itemize}

ANX's security guarantees depend on proper deployment: using a trusted ANX Core, the official ANX Hub, and verifying ANX UI dialog credibility.

\subsection{ANX SOP Runtime and Multi‑Agent Collaboration}

Building on ANX's core security and architecture, ANX SOP emerges as a pivotal application of ANX Markup for task scheduling and execution, specifically designed to enable efficient multi‑agent coordination. At its core, ANX SOP addresses a fundamental challenge in agent systems: natural language ambiguity in SOP transmission. Traditional unstructured natural language SOPs often lead to inconsistent task intent understanding among agents, resulting in low execution accuracy and poor stability. In contrast, ANX SOP leverages schema‑based ANX Markup to explicitly define SOP steps, conditional logic, operation standards, and field constraints, ensuring agents receive unambiguous task instructions.

\textbf{Key Insights from Prior Work on SOPs}

ANX's approach is validated by a landmark study by Grab (2025), which demonstrated that explicitly modeling SOPs as tree structures — using a depth‑first search algorithm to navigate the overall workflow while limiting LLMs to executing individual actions — delivers exceptional efficiency and reliability in high‑risk enterprise scenarios. Their results were particularly compelling:

\begin{itemize}
\item 99.8\% accuracy on account takeover investigation tasks
\item Reduced average processing time from 23 minutes to just 3 minutes
\item Automated 87\% of cases
\end{itemize}

The ANX protocol inherently incorporates this proven approach, with ANX Core parsing structured SOPs and driving the workflow. Building upon this foundation, ANX further extends capabilities through its unified protocol and 3EX architecture innovations, enabling longer‑horizon SOP scheduling, sophisticated multi‑agent collaboration, and seamless agent–human interaction.

Moreover, ANX addresses a critical gap identified by Yuan et al. (2026) in current agent communication protocols: while transport and syntactic layers have matured, protocol‑level semantic mechanisms remain severely limited, forcing semantic responsibilities into prompts and wrappers. ANX resolves this imbalance through the structured semantics of ANX Markup, achieving protocol‑level semantic alignment and eliminating ambiguity.

\begin{figure}[htbp]
  \centering
  \includegraphics{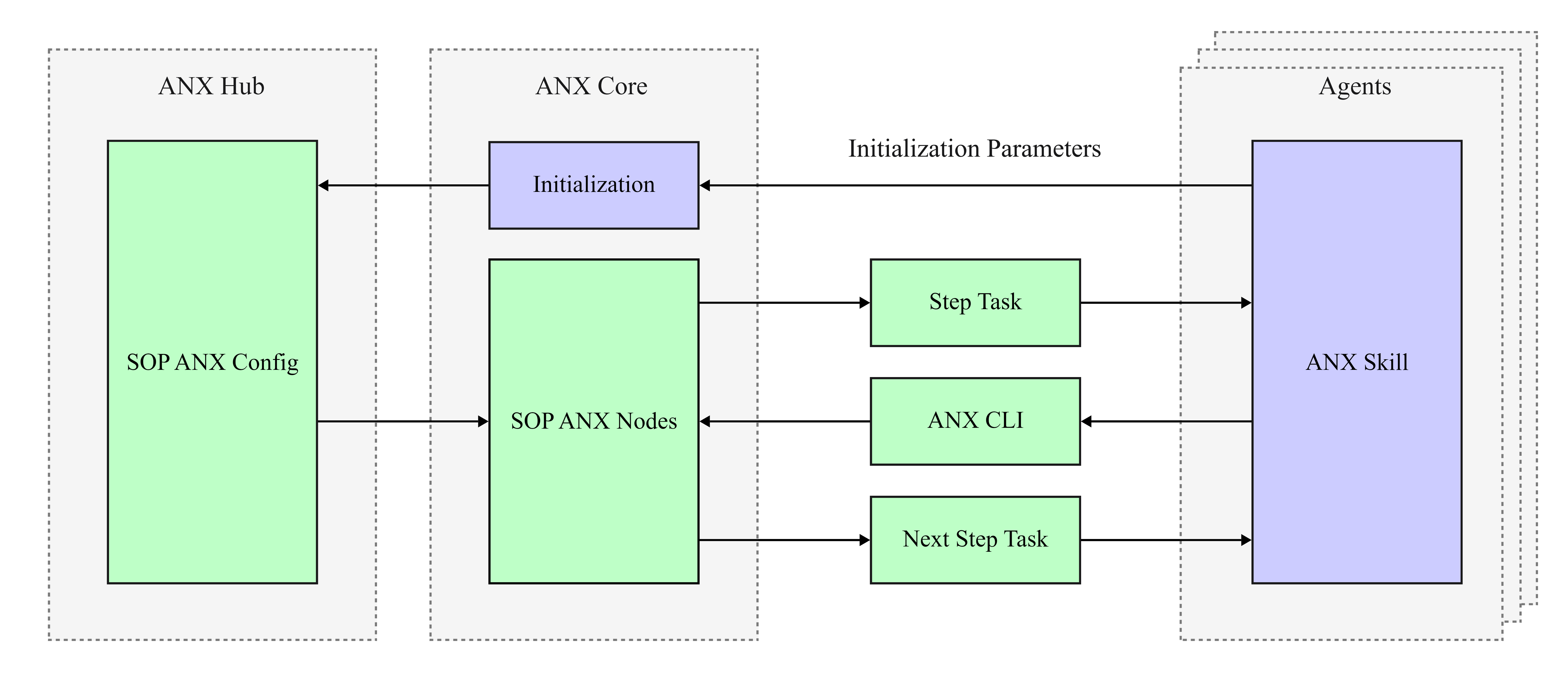}
  \caption{ANX SOP Mechanism}
  \label{fig:sop-mechanism}
\end{figure}

\textbf{Runtime Scheduling.} ANX SOP's scheduling leverages the synergistic interaction of the 3EX layers:

\begin{itemize}
\item
  \textbf{The Expression layer} (ANX Markup) serves as the SOP definition carrier, supporting static step specification, dynamic conditional branch configuration, and step dependency setting to adapt to complex scenarios.
\item
  \textbf{The Exchange layer} (ANXHub) implements dynamic semantic tool discovery over the marketplace and undertakes task routing and cross‑agent synchronization to decompose complex SOPs into parallel sub‑tasks and coordinate agent execution.
\item
  \textbf{The Execution layer} (ANX Core) converts ANX Markup‑structured SOPs into platform‑agnostic ANX CLI commands, adopting progressive disclosure to generate only current‑step execution commands for token efficiency.
\end{itemize}

\textbf{Security Inheritance.} ANX SOP naturally inherits ANX's built‑in security mechanisms: for sensitive data in SOP steps, the core UI‑Core direct communication mechanism (bypassing LLM) ensures robust data isolation, while critical steps trigger human‑only confirmation dialogs to prevent automated misuse.

\textbf{Multi‑Agent Collaboration.} For multi‑agent coordination, ANX Markup provides a unified semantic specification for interoperability, while ANXHub decomposes SOPs based on agent capabilities, assigns specialized sub‑tasks, and synchronizes execution status in real time. This enables ANX SOP to handle large‑scale complex SOPs beyond the capacity of single agents, significantly enhancing both scalability and execution efficiency.

\subsubsection{Case Study: Multi‑Agent and Human Collaboration for Resume Screening}

We illustrate ANX SOP's support for structured multi‑agent and human collaboration using a resume screening example. The workflow involves three roles: a Review Agent, an Action Agent, and an HR human.

\textbf{\textcircled{1} ANX Core computes a matching score (s1)}

The system reads the resume and outputs a matching score score. Automatic routing is performed based on the score:

\begin{itemize}
\item
  \textbf{score \textgreater{} 80 →} directly proceed to s4 (Schedule Interview), executed by the Action Agent.
\item
  \textbf{score \textless{} 60 →} directly proceed to s3 (Politely Decline), executed by the Action Agent.
\item
  \textbf{$60 \leq$ score $< 80$ →} proceed to s2 (Manual Review), triggering a human‑agent collaborative process.
\end{itemize}

\textbf{\textcircled{2} Manual review (s2). For $60 \leq$ score $< 80$:}

\begin{itemize}
\item
  The Review Agent performs supplementary analysis (e.g., skill match) and writes results into an ANX Node.
\item
  The system sends an OA approval request to the HR human via ANX UI, showing resume data, analysis, and "pass"/"reject" options.
\item
  The workflow blocks until HR chooses:
  \begin{itemize}
  \item
    \textbf{pass} → trigger s4 (Action Agent schedules interview)
  \item
    \textbf{reject} → trigger s3 (Action Agent sends decline)
  \end{itemize}
\item
  Agents cannot simulate or bypass the human step.
\end{itemize}

\textbf{\textcircled{3} Centralised state management.} All collaboration information (score, routing target, analysis results, approval status) is managed by ANX Nodes inside ANX Core. Each agent reads inputs from and writes outputs to its assigned Node. HR actions update the Node via ANX UI. Thus, roles communicate indirectly through ANX Core, not directly.

\begin{figure}[htbp]
  \centering
  \includegraphics{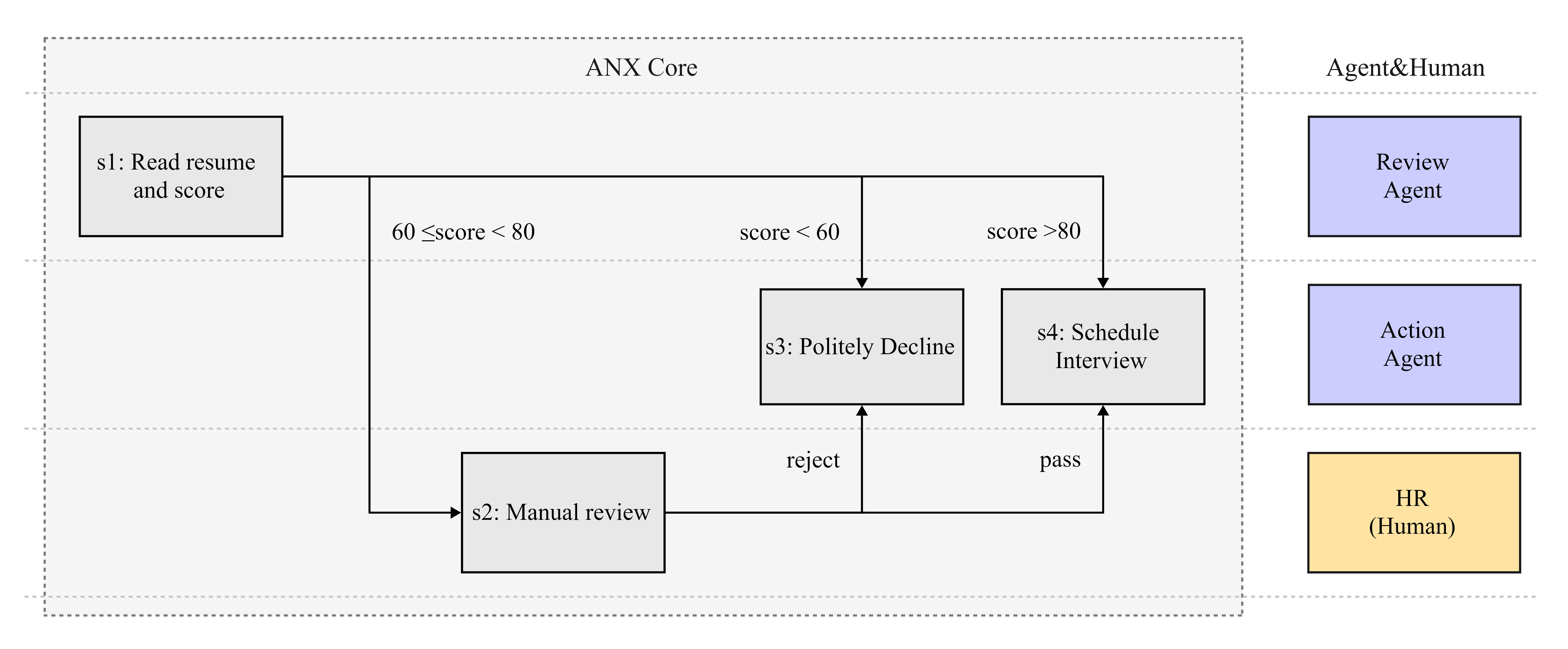}
  \caption{Human \& Multi-Agent Collaborative ANX SOP Mechanism}
  \label{fig:collaboration}
\end{figure}

\textbf{Key aspects.}

\begin{itemize}
\item
  \textbf{Agent‑Agent}: Review Agent and Action Agent share state via ANX Nodes; the former's output feeds the HR interface, the latter executes final decisions.
\item
  \textbf{Agent‑Human}: The 60--80 range forces a blocking approval step, keeping boundary cases under human control. The UI is human‑only; workflow waits for HR action.
\end{itemize}

\textbf{Clarity}: All conditions and target roles are explicitly defined in ANX Config, eliminating natural language ambiguity.

Thus, ANX SOP enables multiple agents and a human to work under a unified workflow: agents handle high‑confidence tasks (auto‑interview, auto‑decline), humans handle boundary cases, and ANX Nodes ensure consistent, traceable state management.

\subsection{Discussion}

\begin{table}[htbp]
  \centering
  \caption{Comparison of GUI‑based Agents, MCP and ANX}
  \label{tab:comparison}
  \fontsize{9}{11}\selectfont
  \begin{tabularx}{\linewidth}{@{}p{0.22\linewidth}p{0.20\linewidth}p{0.20\linewidth}p{0.30\linewidth}@{}}
  \toprule
  Dimension & GUI-Based & MCP & ANX \\
  \midrule
  Architecture & Monolithic & Tool-centric & 3EX Decoupled \\
  Token (Task-Inc, avg) & 8.7k & 6.85k & 3.35k \\
  New Tool Discovery \& Use‑and‑Go & Visual, no & Pre‑load all, no & Dynamic via ANXHub, yes \\
  Human‑Agent Co‑Use & Yes (agent mimics) & Agent‑only & Native shared \\
  Sensitive Data Protection & None (agent sees all) & None & UI‑to‑Core, agent never sees \\
  Human Confirmation & None & None & Human‑only, no programmatic exit \\
  Semantic Precision & Low (NL) & Low--Medium (JSON+MD) & High (structured ANX Markup) \\
  SOP \& Multi‑Agent & Not supported & Limited & Built‑in (long‑horizon + collaboration) \\
  \bottomrule
  \end{tabularx}
\end{table}

*Averaged across Qwen3.5‑plus and GPT‑4o results from Tables 10 and 11.*

The results clearly show that ANX's combination of a structured protocol (ANX Markup, ANX Config, ANX CLI, dynamic discovery) and a decoupled architecture (3EX) achieves superior token efficiency, security, and SOP adaptability compared to both GUI‑based automation and MCP. The architecture's separation of concerns enables progressive disclosure and fixed baseline token consumption, while the protocol's semantic precision eliminates natural language ambiguity. The built‑in security mechanisms (UI‑to‑Core and human‑only confirmation) address fundamental privacy gaps that existing paradigms leave open. Together, these innovations provide a holistic foundation for agent‑native interaction.
\section{Experimental Methodology}

\subsection{Research Questions}

\textbf{RQ1 (Efficiency)} -- Does ANX reduce token consumption and execution time compared to GUI-based automation and MCP-based Skill on a form‑filling task?

\textbf{RQ2 (Accuracy)} -- Does ANX's structured semantics improve task completion accuracy compared to the two baselines?

(Security evaluation and multi‑step SOP evaluation are left as future work.)

\subsection{Task Design}

We evaluate all three methods (GUI, MCP-based Skill, and ANX) on a single representative task: job account registration form filling in a realistic business scenario.

Unlike MCP-Universe (Luo et al., 2025) which provides a broad benchmark across diverse MCP servers, our experiment is designed to evaluate MCP's limitations in a specific real-world complex scenario (i.e., dynamic form filling with remote option loading).

\begin{table}[htbp]
  \centering
  \caption{Experimental Task}
  \label{tab:task}
  \fontsize{9}{11}\selectfont
  \begin{tabularx}{\linewidth}{@{}p{0.25\linewidth}p{0.50\linewidth}p{0.20\linewidth}@{}}
  \toprule
  Task & Description & Complexity \\
  \midrule
  Form Filling & 10‑field job form with two dynamic option fields (e.g., industry) loaded via URL. Sensitive handling disabled. & Medium \\
  \bottomrule
  \end{tabularx}
\end{table}

\emph{No multi‑skill invocation, SOP workflow, or sensitive data handling is involved in this experiment.}

\subsection{Experimental Setup}

\textbf{Framework:} OpenClaw Agent (Dockerized). A baseline of 13.2k tokens is consumed by system prompts and core skills before any task‑specific execution.

\textbf{Models:}

\begin{itemize}
\item
  Qwen3.5‑plus (1M context, temperature 0.7)
\item
  GPT‑4o (128K context, temperature 0.7)
\end{itemize}

\begin{table}[htbp]
  \centering
  \caption{Three compared methods (same business scenario)}
  \label{tab:methods}
  \fontsize{9}{11}\selectfont
  \begin{tabularx}{\linewidth}{@{}p{0.15\linewidth}p{0.22\linewidth}p{0.55\linewidth}@{}}
  \toprule
  Method & Backend / Skill Used & Description \\
  \midrule
  GUI-based Automation & OpenClaw Agent built‑in Playwright + Chromium & The agent visually parses the rendered web page, extracts DOM elements, and simulates human inputs (keyboard/mouse) to fill and submit the form. No custom skill is used. \\
  MCP-based Skill & Model Context Protocol with Job Form Skill & The agent uses declarative field definitions. Dynamic options are pre‑fetched and exposed as tool inputs. The agent calls the job\_form tool with a JSON payload of 10 fields. The tool executes and returns a confirmation. \\
  ANX-based Skill & ANX protocol + 3EX architecture (ANX Markup, ANX CLI, ANX Core, ANXHub) & The agent generates an ANX Markup form definition. ANX Core compiles it into an anx CLI command. Dynamic options are resolved via the URL specified in the markup. Execution runs on ANX Node. \\
  \bottomrule
  \end{tabularx}
\end{table}

All runs are executed in isolated Docker containers with fully reset context (30 repetitions per method per model).

The design of both MCP-based and ANX-based skills follows the token-efficient skill representation principles of SkillReducer (Gao et al., 2026).

\begin{figure}[htbp]
  \centering
  \setlength{\tabcolsep}{12pt}
  \begin{tabular}{cc}
    \includegraphics[width=0.45\linewidth]{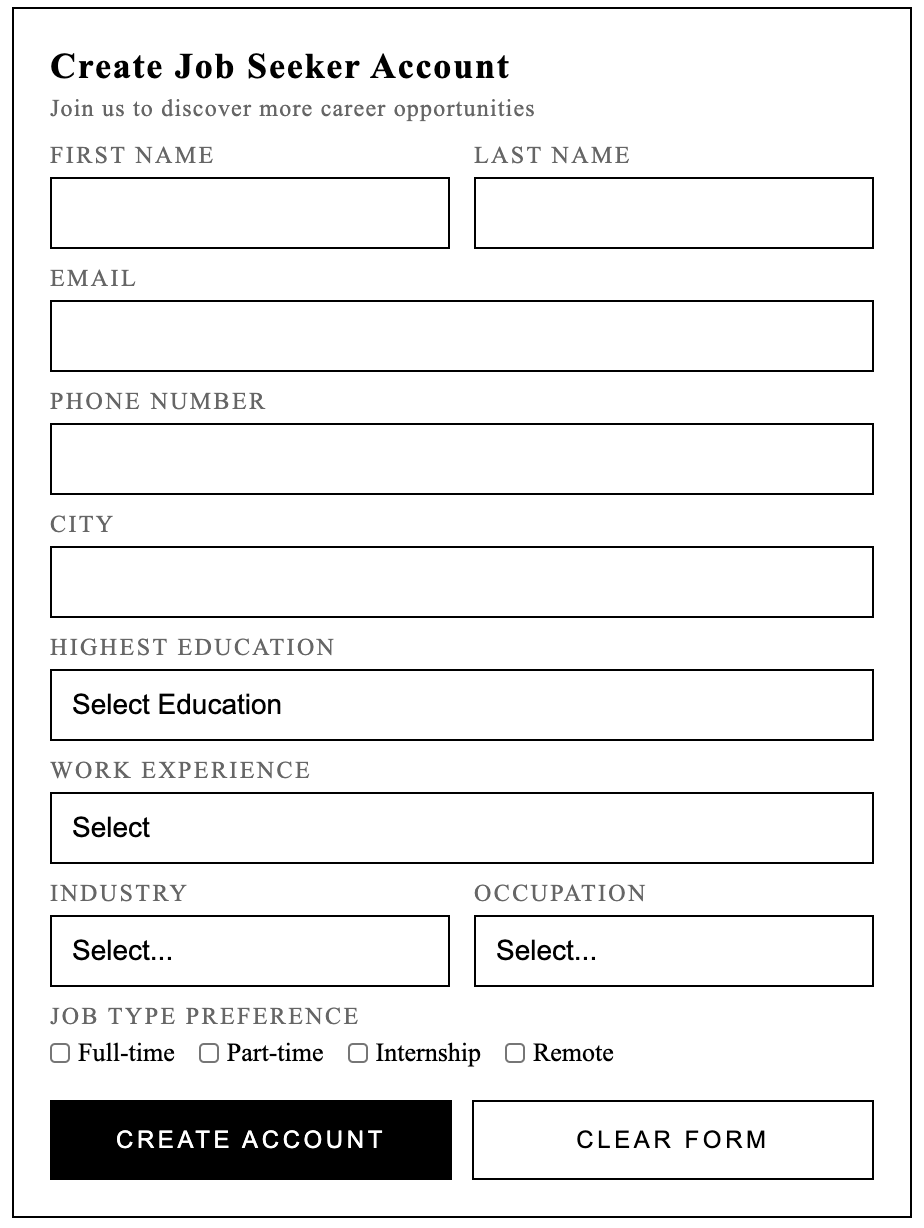} & 
    \includegraphics[width=0.45\linewidth]{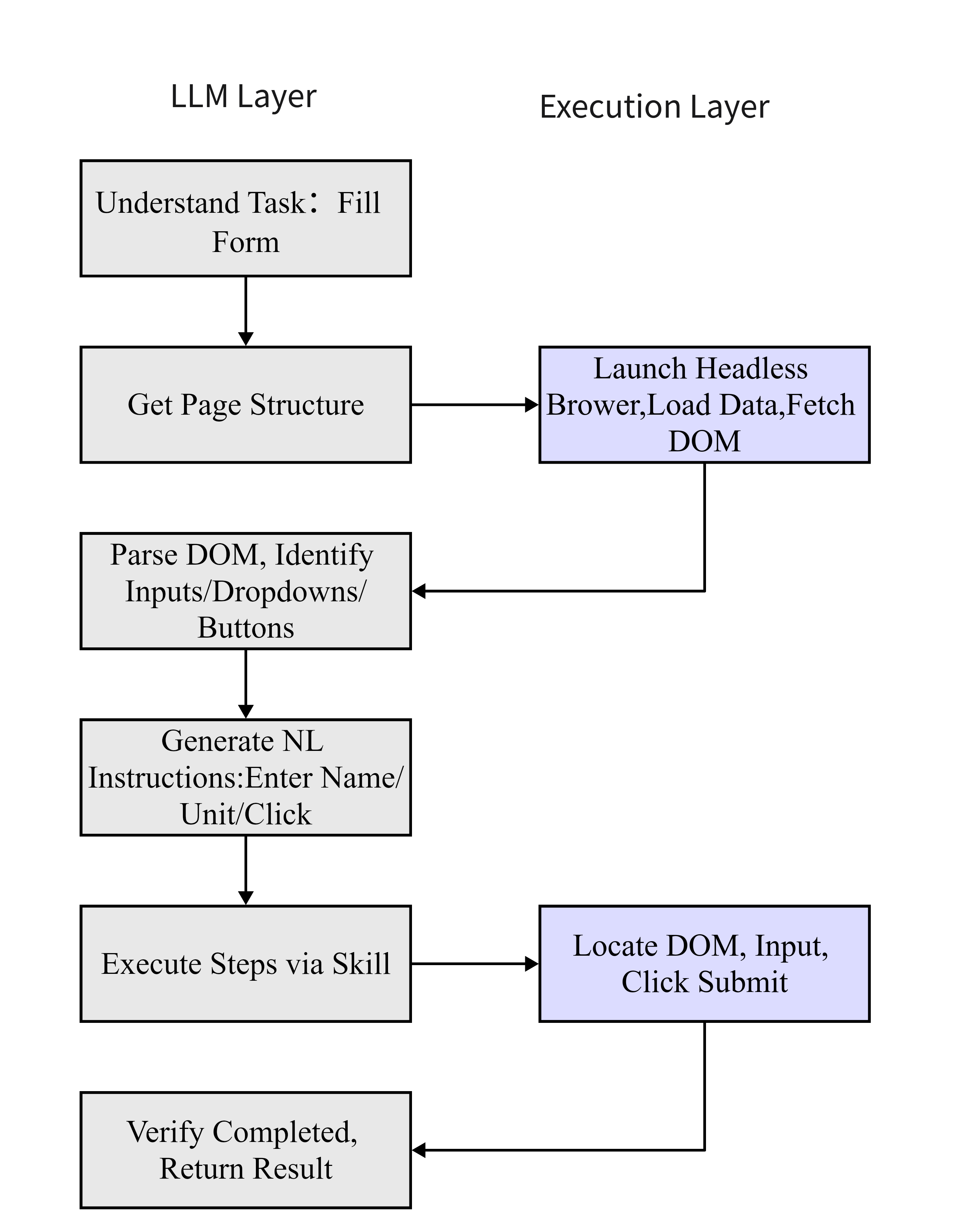} \\
    Figure 7: Create Job Account Form Example & 
    Figure 8: GUI Flow \\[12pt]
    \includegraphics[width=0.45\linewidth]{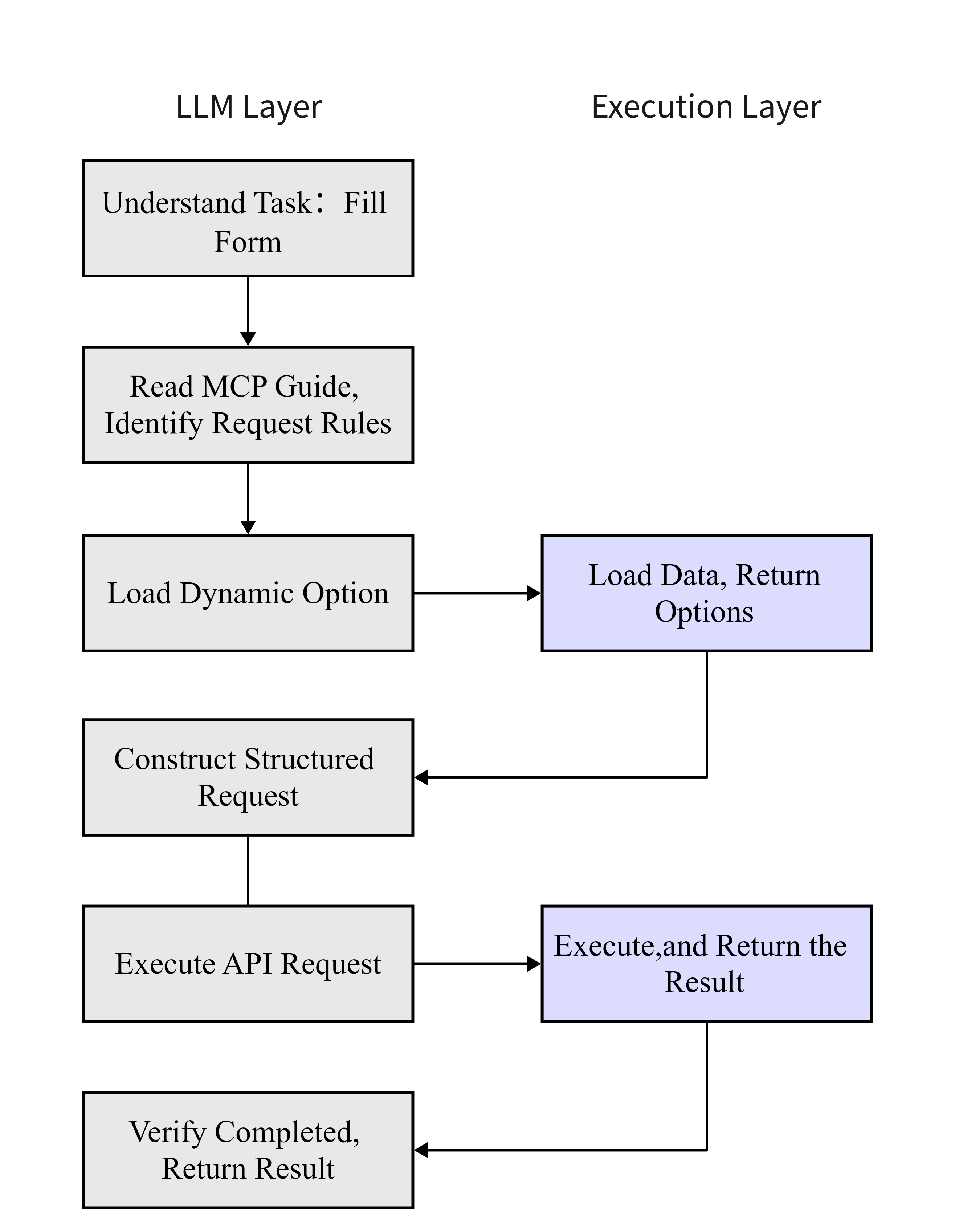} & 
    \includegraphics[width=0.45\linewidth]{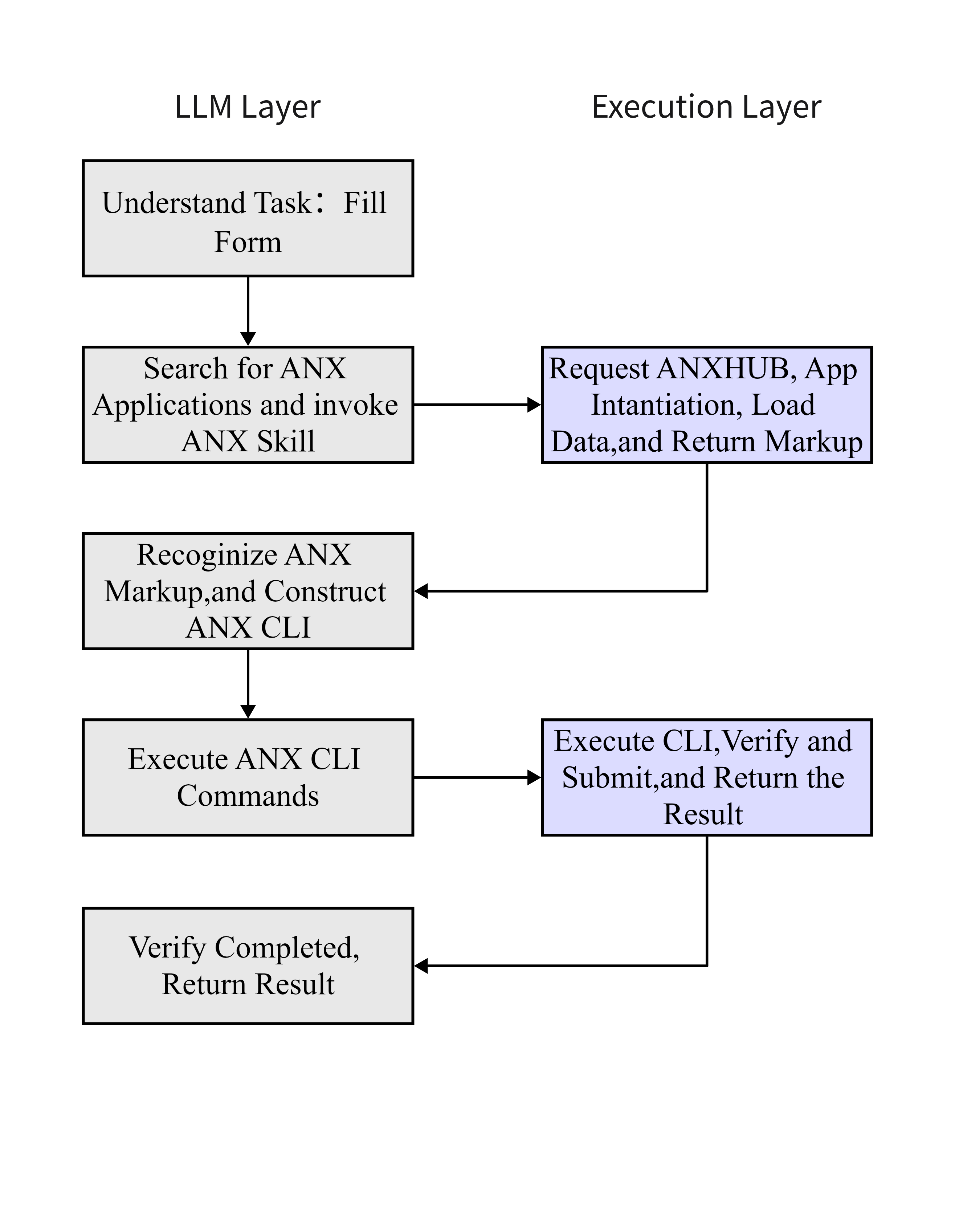} \\
    Figure 9: MCP Flow & 
    Figure 10: ANX Flow \\
  \end{tabular}
  \label{fig:workflows}
\end{figure}

\subsection{Evaluation Metrics \& Statistical Analysis}

\textbf{Metrics:}

\begin{itemize}
\item
  Total Token Consumption
\item
  Task‑Incremental Token (total minus 13.2k baseline)
\item
  Execution Time (end‑to‑end, seconds)
\item
  Task Accuracy (successful completion rate)
\end{itemize}

\textbf{Statistical validation:} Independent two‑sample t‑tests ($\alpha$ = 0.05) with Bonferroni correction, applied separately for each model. For each method and model, we repeat the experiment 30 times (n = 30). Each repetition runs in a fresh Docker container with no shared state. After Bonferroni correction (3 comparisons: GUI vs. MCP, GUI vs. ANX, MCP vs. ANX), only p \textless{} 0.0167 is considered statistically significant.

\subsection{Experimental Case and Procedure}

To empirically evaluate the performance of ANX against GUI-based agents and MCP across the designed task, this section details the experimental case, workflow visualization, and standardized execution procedure. Taking the form filling task as the representative experimental case, we illustrate the implementation details of each method, while providing visual comparisons of their workflows. The unified procedure was strictly followed for all tasks, models, and methods to ensure experimental validity and reproducibility.

\textbf{Procedure:}

1. \textbf{Setup}: Initialize Docker container with fresh context
2. \textbf{Baseline}: Load system prompts and core skills (13.2k tokens)
3. \textbf{Task Execution}: Run the form filling task using the specified method
4. \textbf{Measurement}: Record token consumption, execution time, and task completion status
5. \textbf{Repeat}: Repeat steps 1-4 for 30 iterations per method per model
6. \textbf{Analysis}: Apply statistical tests to compare performance across methods
\section{Results and Discussion}

\subsection{RQ1: Efficiency}

Results for both Qwen3.5-plus and GPT-4o are presented in Tables 10 and 11, respectively, detailing token consumption and execution time.

\begin{table}[htbp]
  \centering
  \caption{Qwen3.5-plus Token Consumption and Execution Time}
  \label{tab:qwen-results}
  \fontsize{9}{11}\selectfont
  \begin{tabularx}{\linewidth}{@{}p{0.16\linewidth}p{0.16\linewidth}p{0.16\linewidth}p{0.20\linewidth}p{0.18\linewidth}@{}}
  \toprule
  Method & Total Token (k) & Baseline (k) & Task‑Inc Token (k) & Execution Time (s) \\
  \midrule
  GUI & 22.3 ± 2.6 & 13.2 & 9.1 ± 2.6 & 33.2 ± 9.5 \\
  MCP-based Skill & 20.6 ± 2.0 & 13.2 & 7.4 ± 2.0 & 30.8 ± 8.2 \\
  ANX & 17.1 ± 1.6 & 13.2 & 3.9 ± 1.6 & 12.9 ± 2.7 \\
  \bottomrule
  \end{tabularx}
\end{table}

\begin{table}[htbp]
  \centering
  \caption{GPT-4o Token Consumption and Execution Time}
  \label{tab:gpt-results}
  \fontsize{9}{11}\selectfont
  \begin{tabularx}{\linewidth}{@{}p{0.16\linewidth}p{0.16\linewidth}p{0.16\linewidth}p{0.20\linewidth}p{0.18\linewidth}@{}}
  \toprule
  Method & Total Token (k) & Baseline (k) & Task‑Inc Token (k) & Execution Time (s) \\
  \midrule
  GUI & 21.5 ± 2.8 & 13.2 & 8.3 ± 2.8 & 48.2 ± 12.2 \\
  MCP-based Skill & 19.5 ± 2.1 & 13.2 & 6.3 ± 2.1 & 39.0 ± 9.1 \\
  ANX & 16.0 ± 1.0 & 13.2 & 2.8 ± 1.0 & 16.5 ± 4.8 \\
  \bottomrule
  \end{tabularx}
\end{table}

For Qwen3.5-plus, the reduction in task-incremental tokens with ANX compared to MCP-based Skill was 3.5k (95\% CI: [2.56k, 4.44k], t(55.3)=7.48, $p < 0.001$), and execution time was reduced by 17.9s (95\% CI: [14.71s, 21.09s], t(38.6)=11.36, $p < 0.001$). For GPT-4o, the reduction in task-incremental tokens was 3.5k (95\% CI: [2.645k, 4.355k], t(44.6)=8.24, $p < 0.001$), and execution time was reduced by 22.5s (95\% CI: [18.72s, 26.28s], t(45.9)=11.98, $p < 0.001$). All comparisons remained significant after Bonferroni correction ($\alpha$=0.0167).

\subsection{Discussion}

The results in Tables 10 and 11 demonstrate clear efficiency advantages of ANX over both the MCP‑based Skill and GUI automation for the job account registration form‑filling task.

\textbf{Token consumption.} ANX reduces task‑incremental token consumption by 47.3\% (Qwen3.5‑plus) and 55.6\% (GPT‑4o) compared to the MCP‑based Skill, and by 57.1\% and 66.3\% compared to GUI. The primary reason is that ANX delegates form rendering and dynamic option resolution to ANX Core and the CLI layer, rather than forcing the LLM to handle these details. In the GUI baseline, the LLM must process the entire DOM tree and generate low‑level browser actions (clicks, key presses), leading to high token usage. The MCP‑based Skill improves over GUI by using a declarative tool schema, but it still requires the LLM to actively load and pass dynamic option sets (e.g., the industry dropdown from a remote dataset) as part of the tool call payload. ANX avoids this by specifying only a URL in the ANX Markup; ANX Core automatically fetches the options at execution time, eliminating token overhead for dynamic data loading. This advantage is consistent across both LLM models.

\textbf{Execution time.} ANX also achieves the lowest end‑to‑end execution time: 12.9\,s (Qwen) and 16.5\,s (GPT‑4o), compared to 30.8\,s and 39.0\,s for the MCP‑based Skill, and 33.2\,s and 48.2\,s for GUI. The GUI baseline suffers from slow DOM parsing, sequential element interaction, and page load waits. The MCP‑based Skill reduces time by bypassing visual rendering, but still incurs overhead from serializing and transmitting the full JSON payload, including dynamically loaded options. ANX's CLI execution and direct ANX Node invocation minimize serialization overhead and eliminate round‑trips for dynamic option fetching, as the Core handles resolution locally.

\textbf{Generality across models.} The relative efficiency gains are similar for Qwen3.5‑plus and GPT‑4o, indicating that ANX's architecture---not model‑specific optimization---is the driving factor.

\textbf{Accuracy considerations.} While we focused on efficiency metrics in this evaluation, ANX's structured semantics also contribute to improved task completion accuracy. The clear, machine‑executable ANX Markup eliminates natural language ambiguity, reducing the likelihood of execution errors. This is particularly important for complex tasks with multiple steps or conditional logic.

\textbf{Practical implications.} The token efficiency gains translate directly to cost savings, especially for models like GPT‑4o where token usage is a significant cost factor. Additionally, the reduced execution time improves user experience by providing faster task completion, which is crucial for interactive applications.

In summary, for the evaluated form‑filling scenario with dynamic options, ANX's combination of structured markup (ANX Markup), decoupled execution (3EX), and CLI‑based command generation delivers substantial improvements in both token efficiency and runtime. The advantage is rooted in shifting dynamic data resolution from the LLM's context to the execution layer, a design that avoids the pre‑loading and active option handling required by the MCP‑based Skill.

\section{Conclusion and Future Work}

\subsection{Conclusion}

This paper represents the first in the ANX (AI Native eX) series, focusing on introducing the core protocol design and 3EX decoupled architecture. As the inaugural work in this series, it lays the foundation for agent-native interaction by presenting a comprehensive framework that addresses the key challenges of efficiency, security, and collaboration in AI agent systems.

ANX (AI Native eX) is a three-layer protocol architecture that introduces the 3EX (Expression-Exchange-Execution) decoupled framework—separating task specification, tool discovery, and execution into independent layers. This architectural foundation enables three core innovations:

ANX provides a scalable, secure, cost-effective foundation for AI agent deployment in human-machine collaboration scenarios. At this stage, experimental validation is limited to specific task scenarios with a constrained dataset, and further large-scale evaluations are still needed to verify the protocol's stability and efficiency in complex, cross-platform environments.

\subsection{Future Work}

We have proposed and initially demonstrated the feasibility of the ANX framework, and we are excited to share it with the community. However, we acknowledge that more work and more rigorous experiments are needed to fully validate its capabilities. Moving forward, we will focus on the following seven key directions to address these limitations and further advance the ANX protocol:

  • \textbf{Multi-Agent Coordination} – Extend to hybrid protocols (Dochkina, 2026) and incorporate dynamic agent elimination (Wang et al., 2025) to optimize token efficiency and task performance. Our current ANXHub implements task routing primitives; full evaluation with multiple agents is left for future work.

  • \textbf{Sensitive Data and Sensitive Operation Research} – Further investigate ANX's UI-to-Core isolation and human-only confirmation mechanisms through systematic experiments (e.g., adversarial attacks, information leakage tests). Strengthen security by addressing remaining vulnerabilities such as UI spoofing, untrusted Core/Hub, and LLM-induced social engineering. Explore the trade-off between security and usability: evaluate the overhead of human confirmations and sensitive field redaction, and design adaptive policies that reduce user friction while maintaining strong guarantees for high-risk operations.

  • \textbf{ANXHub vs. SkillHub Discovery and Download} – Conduct a systematic comparison between ANXHub (dynamic marketplace, use-and-go, no installation) and traditional SkillHub / MCP-style tool indices (pre-registration, local binaries) across four key dimensions: timeliness (cold-start latency, update propagation), security (supply chain attacks, virus risks, dependency vulnerabilities), convenience (zero-install vs. pre-download, dynamic vs. static discovery), and token consumption (progressive disclosure in ANXHub vs. pre-loaded schemas in SkillHub). Evaluate under realistic network conditions, varying skill library sizes, and concurrent request scenarios to quantify the practical advantages of ANX's Exchange layer.

  • \textbf{Enterprise AI Carrier \& Weakly-Decentralized ANXHub} – Investigate AI Carrier-based enterprise architecture, enabling organizations to deploy their own ANXHub instances with federated trust models, multi-Hub trust policies, and semantic discovery optimization, achieving weak centralization while reducing dependency on a single official Hub.

  • \textbf{Complex and Compute-Intensive Scenarios} – Extend ANX to handle more complex task graphs (e.g., 50+ steps with nested branches) and compute-intensive scenarios (e.g., real-time data processing, large-scale simulations), evaluating the scalability and resource efficiency of the 3EX architecture under high load.

  • \textbf{Skill Growing} – Integrate with skill-growing paradigms (Chen et al., 2026) to enable adaptive skill learning and dynamic schema evolution.

  • \textbf{Traceability} – Develop richer traceability mechanisms (Peng, 2026) to track interaction lifecycles and enable auditability for regulatory compliance.
  
  • \textbf{Lightweight Specialized Models for ANX Protocol} – Following the success of fine-tuned models for API calling (Gorilla, Patil et al., 2024), we plan to train smaller models specialized for ANX protocol tasks (e.g., ANX Markup to CLI compilation), enabling lower-cost deployment and faster inference.

  \subsection{Final Remarks}

We welcome researchers, engineers, and industrial partners to join the development of the ANX protocol, contribute to its iteration, and jointly build a more open, reliable, and unified ecosystem for AI-native applications.

% References
\makeatletter
\renewcommand{\@biblabel}[1]{}
\def\thebibliography#1{\section*{\refname}\@mkboth{\MakeUppercase\refname}{\MakeUppercase\refname}\list{}{\setlength{\labelwidth}{0pt}\setlength{\leftmargin}{0pt}\setlength{\itemindent}{2em}\setlength{\labelsep}{0pt}\usecounter{enumiv}}\sloppy\clubpenalty4000\@clubpenalty \clubpenalty\widowpenalty4000\sfcode`\.\@m}
\def\endthebibliography{\def\@noitemerr{\@latex@warning{Empty `thebibliography' environment}}\endlist}
\makeatother

\section*{Appendix}

\vspace{6pt}

\noindent Full ANX Markup schemas, experimental scripts, and detailed results are maintained in the open-source repository:

\vspace{12pt}

\begin{itemize}
  \item \textbf{Core implementation:} \url{https://github.com/mountorc/anx-core}
  \item \textbf{Protocol specification:} \url{https://github.com/mountorc/anx-protocol}
\end{itemize}

\section*{Acknowledgements}

\noindent We thank Wang Xin for inspiring discussions and constructive feedback on
the conception and terminology refinement of this work.

\end{document}